\documentclass[letterpaper,twocolumn,fleqn]{article} 

\usepackage{ist}
\usepackage{graphicx}
\usepackage{amsmath}
\usepackage{amssymb}
\usepackage{caption, subcaption}
\usepackage{cite}
\usepackage{array}
\usepackage{booktabs}
\usepackage{makecell}

\title{A Color Image Analysis Tool to Help Users Choose a Makeup Foundation Color*}

\author{ Yafei Mao $^a$, Christopher Merkle $^b$, Jan P. Allebach $^a$ \\ $^a$ School of Electrical and Computer Engineering, Purdue University, West Lafayette, IN 47907, U.S.A.\\ $^b$ MIME Inc. Vancouver, WA 98660, U.S.A. }

\date{} 

\begin{document} 

\maketitle 
\let\thefootnote\relax\footnotetext{*Research supported by MIME Inc. Vancouver, WA.}
\thispagestyle{empty} 


\begin{abstract}
This paper presents an approach to predict the color of skin-with-foundation based on a no makeup selfie image and a foundation shade image. Our approach first calibrates the image with the help of the color checker target, and then trains a supervised-learning model to predict the skin color. In the calibration stage, We propose to use three different transformation matrices to map the device dependent $RGB$ response to the reference CIE $XYZ$ space. In so doing, color correction error can be minimized. We then compute the average value of the region of interest in the calibrated images, and feed them to the prediction model. We explored both the linear regression and support vector regression models. Cross-validation results show that both models can accurately make the prediction.
 \end{abstract}

\section{Introduction}
\label{sec:intro}
With the rapid evolution of mobile phone technologies, people can do more and more things on their phones. Virtual makeup try-on is one of the popular mobile applications nowadays. It allows users to test out makeup shades via images or live camera, which makes cosmetics shopping a lot more convenient and fun. Many researches have focused on this field. Bhatti et al \cite{bhatti2010mobile} developed a mobile image-based application to give women personalized cosmetics recommendations. Tong et al \cite{tong2007example} extracted makeup information from before-and-after image pairs and transferred the makeup to a new face. Work in \cite{nguyen2017smart} is based on a similar idea. There are also some papers using deep neural networks \cite{alashkar2017examples, li2015deep} to analyze makeup styles. Most of these papers are targeting facial attribute analysis and style transfer. In this paper, we will study the color change of the skin before-and-after the makeup is applied. We will focus on a specific line of foundation products. 

Taking images with a mobile phone can be easy, but extracting reliable colors from the images remains a problem. Because of imperfect lighting conditions, different camera sensors' sensitivities, and various post-processing in the image processing pipeline of the camera, the same product can look different in different images. 

To address this color disparity issue, researchers have developed various color correction algorithms. These algorithms can be generally classified into two groups: color constancy and image calibration. Color constancy is done by estimating and removing the influence of illuminations. Some popular color constancy algorithms include Gray World and Max RGB. A major limitation of these two algorithms is that they strongly rely on particular assumptions. If some of the assumptions do not hold, then the estimation can be inaccurate. Some more recent neural network algorithms \cite{cheng2014illuminant, barron2017fast, hu2017fc4} can achieve high accuracy but also require lots of computer resources, compared to some traditional methods. On the other hand, image calibration approaches are simple and effective. They directly map the device-dependent $RGB$ color values to some standard color values with the help of a calibration target. This mapping can be applied to any camera and makes very few assumptions about the characteristics of the images taken by the camera. The mapping includes three-dimensional look-up tables combined with interpolation and extrapolation \cite{hung1993colorimetric}, machine learning models \cite{zhao2016facial}, and neural networks \cite{cheung2002color,kang1992neural}. Despite the fact that a wide variety of calibration methods are available, researchers have favored regression-based approaches \cite{berns1995colorimetric,finlayson1997constrained,gindi2008color} due to their simplicity and feasibility.

In this paper, we propose an image analysis tool that can predict the skin-with-foundation color based on the color information retrieved from calibrated selfie and foundation swatch images. To avoid illuminance inconsistency across multiple images of the same subject, we use a protocol to collect image data under controlled lighting conditions. To minimize color correction errors, we group the color patches on the color checker into three sets and compute the mapping from the camera-dependent $RGB$ space to the standard CIE $XYZ$ for these three sets separately. Next, the CIE $L^*a^*b^*$ color coordinates of the skin pixels, foundation pixels, and the skin with foundation pixels are extracted. Finally, a linear regression model and a support vector regression (SVR) \cite{drucker1996support} model are trained using these color coordinates. 


\section{Data Collection}\label{sec: datacollection}
\begin{figure}[t]
\centering
  \centerline{\includegraphics[width=0.38\textwidth]{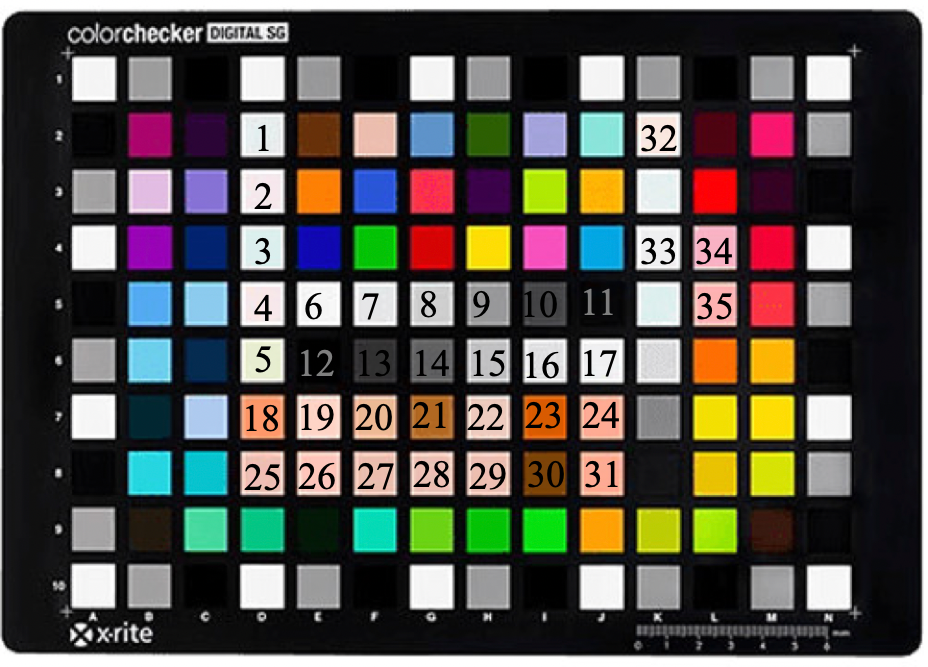}}
  \caption{X-rite digital SG 140 color checker. Patches that are used in the color correction stage are numbered from 1 to 35. }
  \label{fig: cc}
\end{figure}

\begin{figure}[t]
\centering
\begin{subfigure}[b]{0.265\textwidth}
  \centering
  \includegraphics[width=\textwidth]{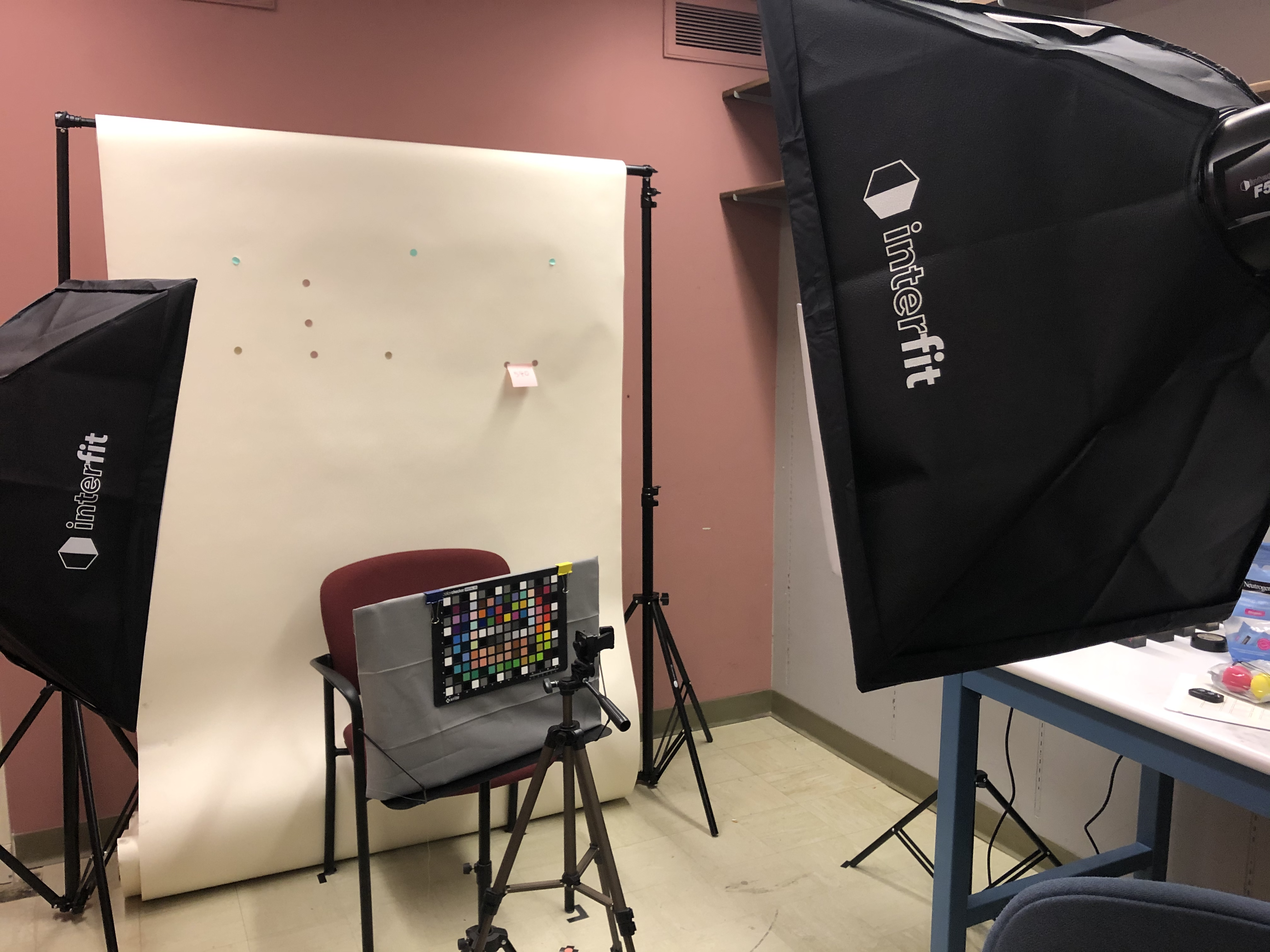}
  \centering
  \caption{\label{subfig: fig1}}
\end{subfigure}
\begin{subfigure}[b]{0.181\textwidth}
  \centering
  \includegraphics[width=\textwidth]{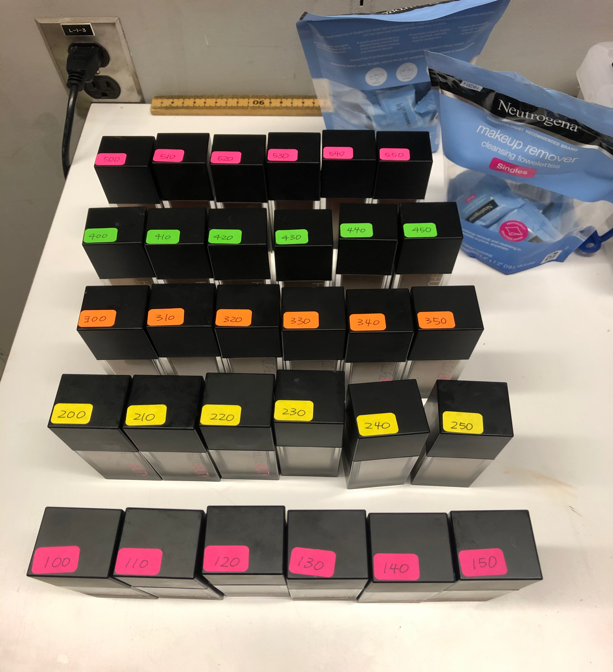}
  \centering
  \caption{\label{subfig: fig2}}
\end{subfigure}
  \caption{Data collection experiment (a) lab setting and (b) the makeup foundation bottles. }
  \label{fig: lab}
\end{figure}

The image calibration method proposed in this paper relies on a standard calibration target. We used the X-rite ColorChecker digital SG 140, which is shown in Fig. \ref{fig: cc}. The target foundation products are shown in Fig. \ref{fig: lab} (b). The detailed procedure is as the following:
\begin{itemize}
\item Take a selfie photo with no foundation applied. 
\item Choose 3 or 4 foundation shades that are close to the actual skin tone by testing each possible match on the skin. 
\item Evenly apply the foundation on the skin using a new and clean sponge. 
\item Wait about three minutes for the foundation to dry, and then take a selfie photo.
\item Remove the foundation completely using makeup removal wipes. 
\item Wait about three minutes for the skin to calm down, and apply the next shade. 
\item Repeat the process until all chosen shades are applied and photographed.
\end{itemize}

\begin{figure}[t]
\centering
\begin{subfigure}[b]{0.11\textwidth}
  \centering
  \includegraphics[width=\textwidth]{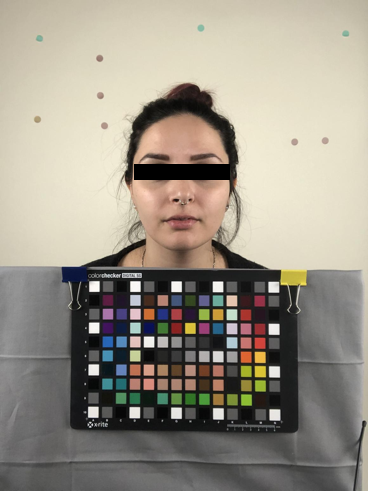}
  \centering
  \caption{\label{subfig: fig1}}
\end{subfigure}
\begin{subfigure}[b]{0.11\textwidth}
  \centering
  \includegraphics[width=\textwidth]{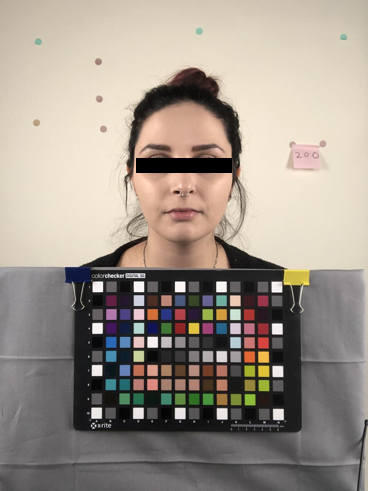}
  \centering
  \caption{\label{subfig: fig2}}
\end{subfigure}
\begin{subfigure}[b]{0.11\textwidth}
  \centering
  \includegraphics[width=\textwidth]{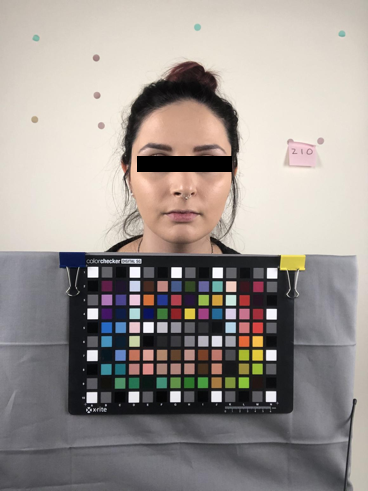}
  \centering
  \caption{\label{subfig: fig3}}
\end{subfigure}
\begin{subfigure}[b]{0.11\textwidth}
  \centering
  \includegraphics[width=\textwidth]{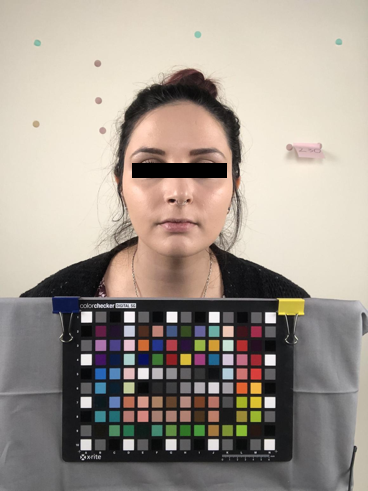}
  \centering
  \caption{\label{subfig: fig4}}
\end{subfigure}
\begin{subfigure}[b]{0.11\textwidth}
  \centering
  \includegraphics[width=\textwidth]{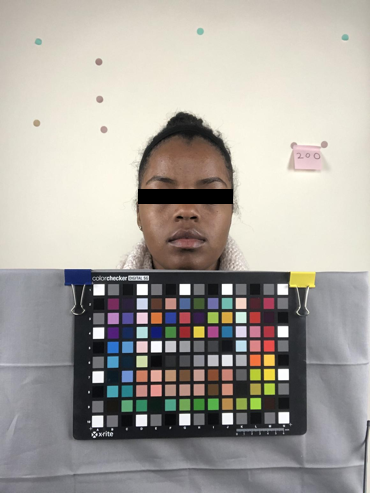}
  \centering
  \caption{\label{subfig: fig1}}
\end{subfigure}
\begin{subfigure}[b]{0.11\textwidth}
  \centering
  \includegraphics[width=\textwidth]{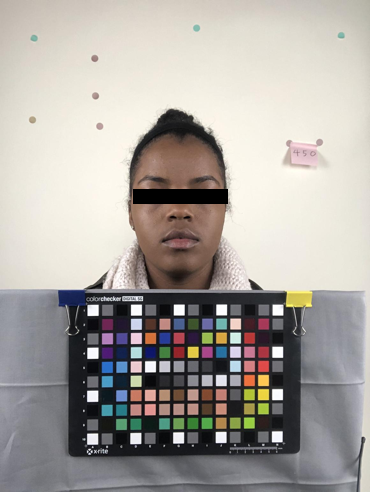}
  \centering
  \caption{\label{subfig: fig2}}
\end{subfigure}
\begin{subfigure}[b]{0.11\textwidth}
  \centering
  \includegraphics[width=\textwidth]{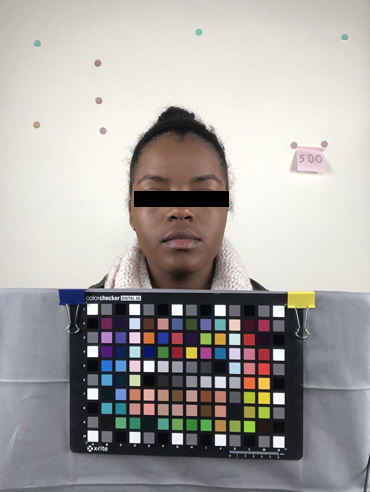}
  \centering
  \caption{\label{subfig: fig3}}
\end{subfigure}
\begin{subfigure}[b]{0.11\textwidth}
  \centering
  \includegraphics[width=\textwidth]{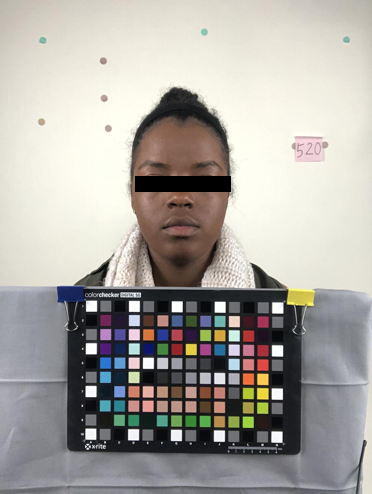}
  \centering
  \caption{\label{subfig: fig4}}
  \end{subfigure}
\caption{ Sample original images. Subject No.1 (a light-skinned subject): (a) skin with no foundation, (b) skin with foundation shade No. 130, (c) skin with foundation shade No. 150, and (d) skin with foundation shade No. 200. Subject No. 2 (a dark-skinned subject): (a) skin with no foundation, (b) skin with foundation shade No. 450, (c) skin with foundation shade No. 500, and (d) skin with foundation shade No. 520. }
\label{fig: originalimages}
\end{figure}

The experiment was conducted under controlled lighting in a lab. The lab setting is shown in Fig. \ref{fig: lab} (a). The light sources were three 4700K LED light bulbs. We installed diffuser panels to make the light as well-diffused as possible. The color checker and the mobile phone are mounted on tripods. The subject used a Bluetooth remote button to take selfie photos. Each subject was instructed to sit at a fixed distance from the camera, so that the amount of the light reflected from his or her face is about the same for all images. The images in Fig. \ref{fig: originalimages} are sample original photos that we collected during this experiment.

\begin{figure}[t]
\centering
\begin{subfigure}[b]{0.12\textwidth}
  \centering
  \includegraphics[width=\textwidth]{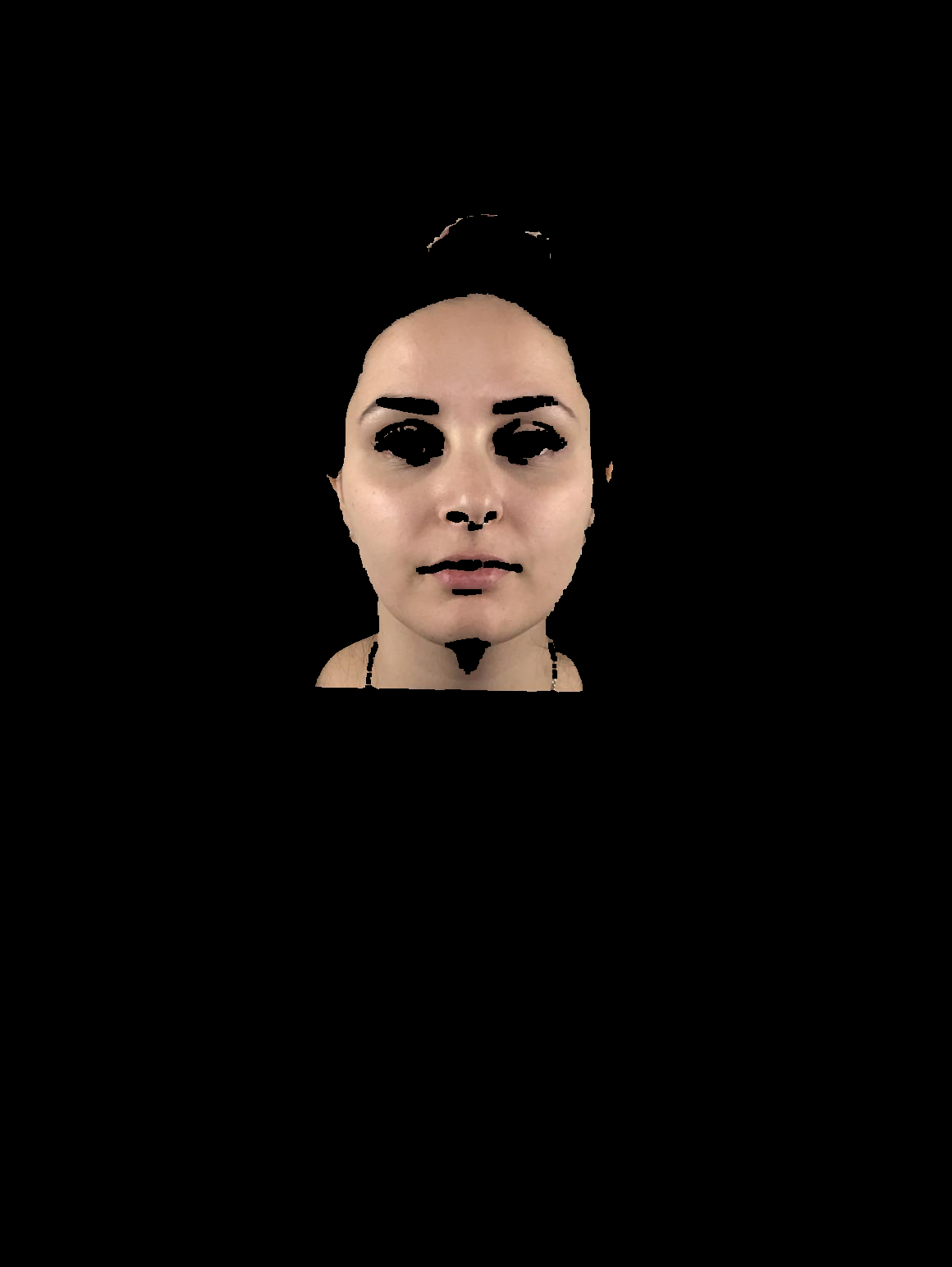}
  \centering
  \caption{\label{subfig: fig2}}
\end{subfigure}
\quad
\begin{subfigure}[b]{0.12\textwidth}
  \centering
  \includegraphics[width=\textwidth]{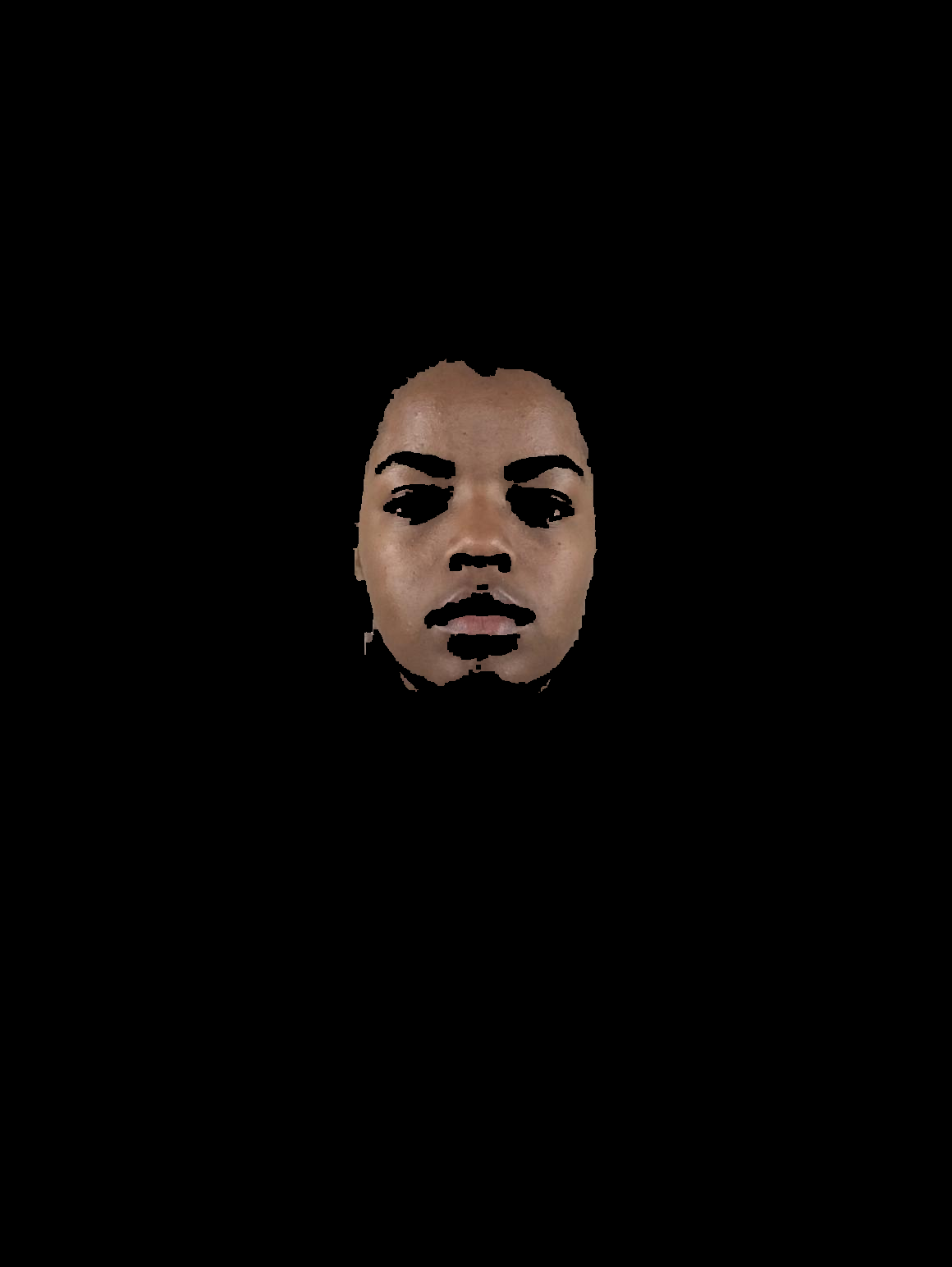}
  \centering
  \caption{\label{subfig: fig4}}
\end{subfigure}
\quad
\begin{subfigure}[b]{0.12\textwidth}
  \centering
  \includegraphics[width=\textwidth]{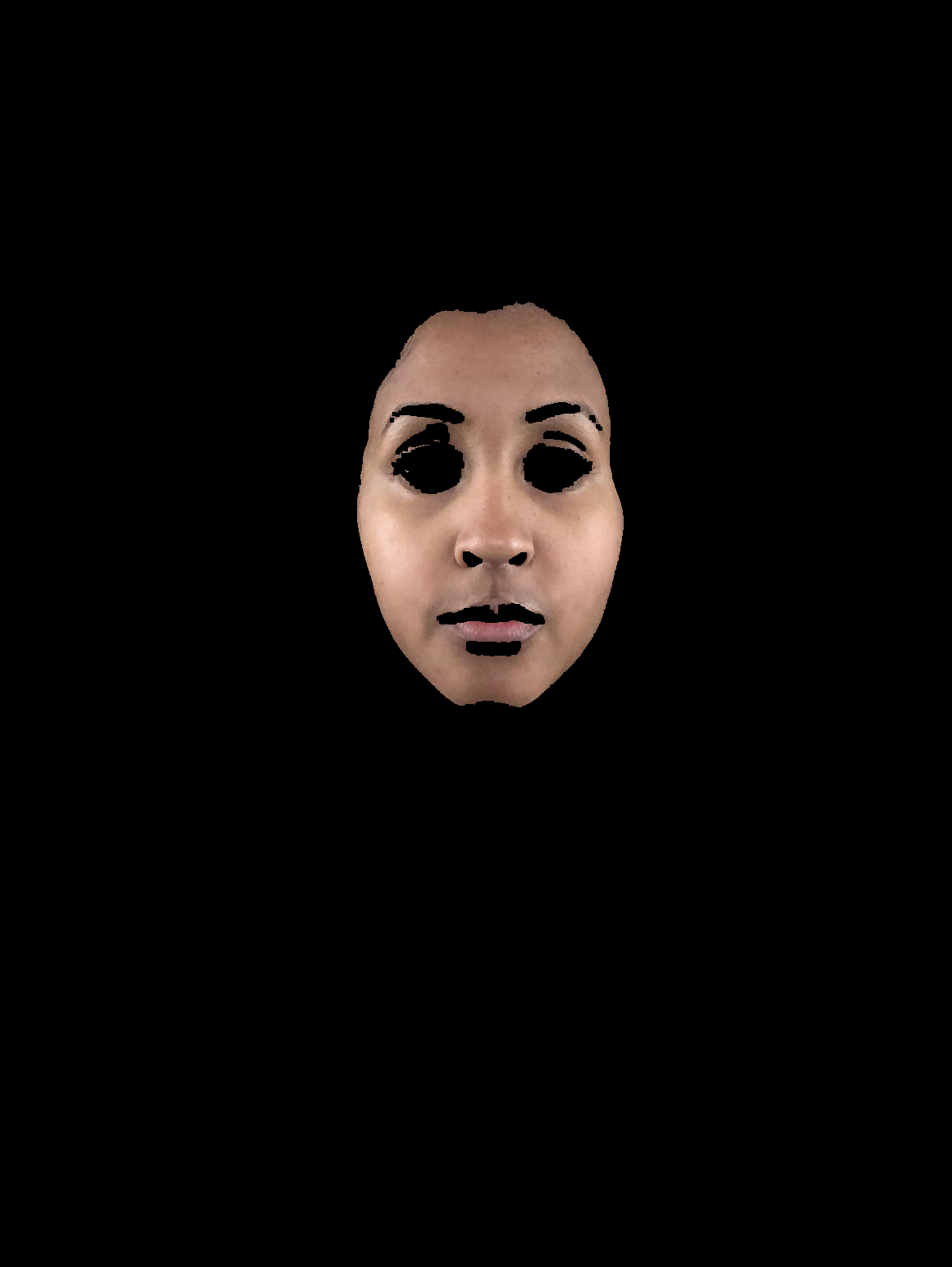}
  \centering
  \caption{\label{subfig: fig4}}
\end{subfigure}
\caption{Skin detection output of the skin with no foundation images of (a) subject No.1 (light-skinned), (b) subject No. 2 (dark-skinned), and (c) subject No. 3 (tan-skinned).}
\label{fig: skindetection}
\end{figure}

\begin{figure}[t]
  \centering
\begin{subfigure}[b]{0.25\textwidth}
  \centering
  \includegraphics[width=\textwidth]{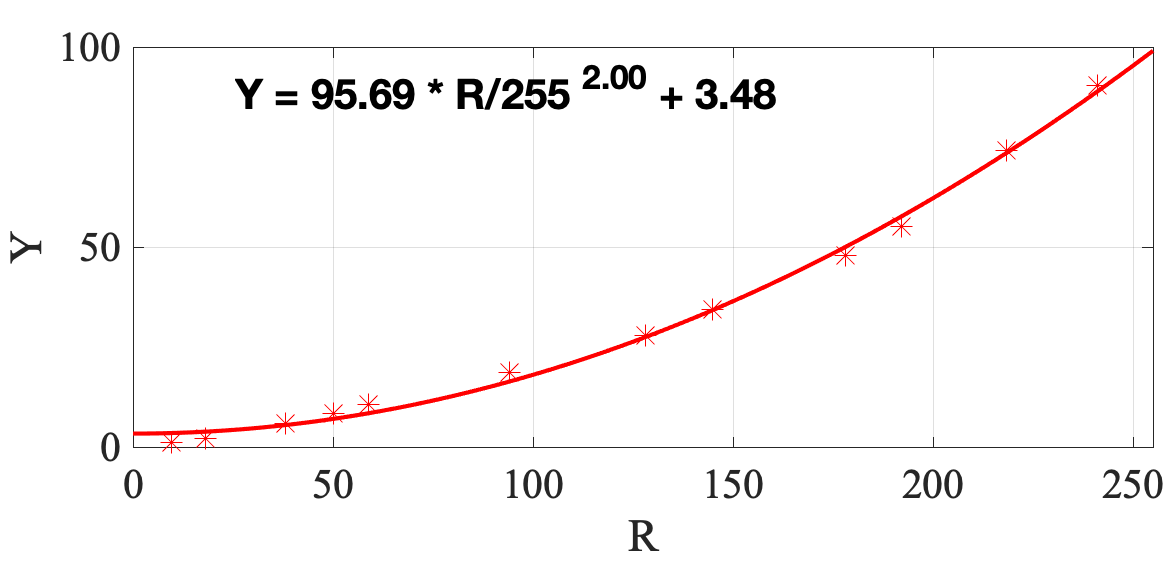}
  \centering
  \caption{\label{subfig: fig1}}
\end{subfigure}
\begin{subfigure}[b]{0.25\textwidth}
  \centering
  \includegraphics[width=\textwidth]{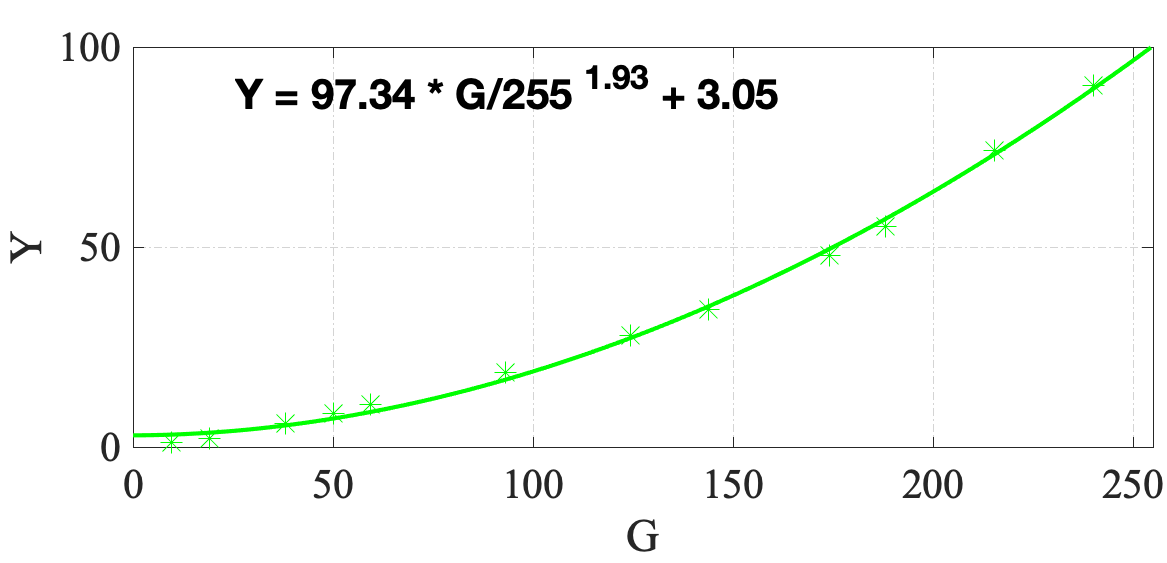}
  \centering
  \caption{\label{subfig: fig2}}
\end{subfigure}
\begin{subfigure}[b]{0.25\textwidth}
  \centering
  \includegraphics[width=\textwidth]{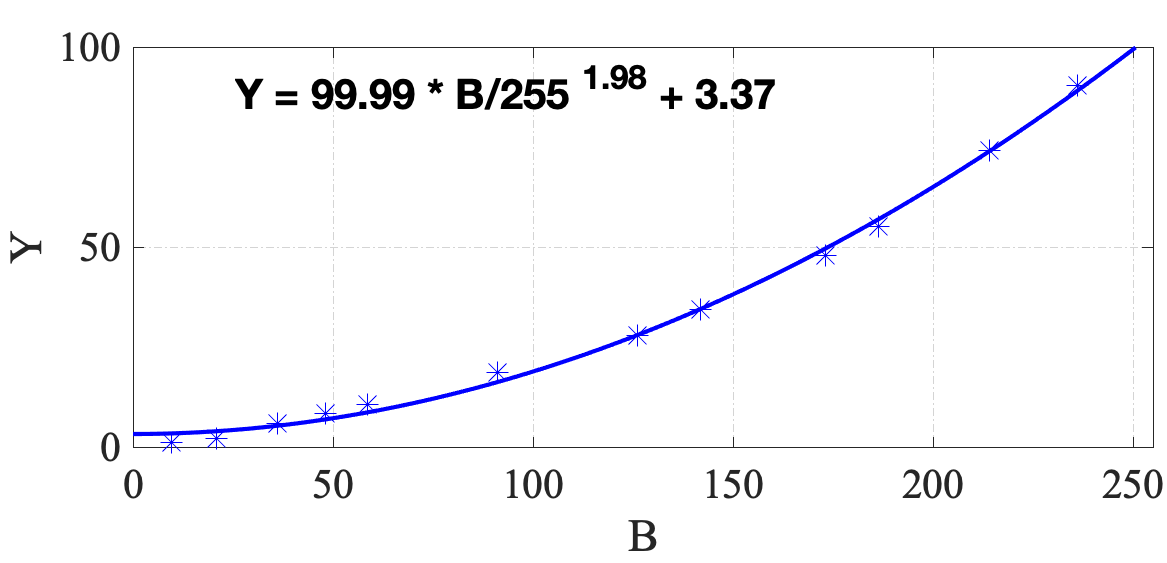}
  \centering
  \caption{\label{subfig: fig2}}
\end{subfigure}
\caption{ The gray balancing curves of the skin with no foundation image of subject No. 1. (a) The $R$ channel. (b) The $G$ channel. (c) The $B$ channel. }
\label{fig: graycurves}
\end{figure}

\section{Image Calibration}
In this section, we will present the details of the image calibration procedure, which will lay the groundwork for an accurate prediction model. We will start by introducing the skin pixel detection algorithm. 
And then we describe the two major steps in the proposed calibration framework. Finally, the calibration performance is evaluated by means of the color differences in CIE $L^*a^*b^*$ space.

\subsection{Skin Detection}\label{sec: skindection}
Since we are mainly interested in analyzing and predicting skin colors, the facial skin is first segmented. A fast and efficient RGB-H-CbCr model \cite{rahman2006rgb} for this purpose is adopted. Under a uniform illumination condition, a skin pixel, defined by \cite{rahman2006rgb}, should satisfy all of the following three criteria. \\
Criterion 1:
\begin{equation}
 \begin{aligned}
R > 95  \quad \land \quad G &> 40 \quad \land \quad B > 20 \\
\max (R,G,B) &-  \min (R,G,B) > 15 \\
|R - G| > 15 \quad \land \quad R &>  G \quad \land \quad R >  B \\
\end{aligned}
\end{equation} 
Criterion 2:
\begin{equation}
 \begin{aligned}
C_r \leq 1.5862 \cdot C_b + 20 \quad \land\\
C_r \geq 0.3448 \cdot C_b + 76.2069 \quad \land\\
C_r \geq -4.5652 \cdot C_b + 234.5652 \quad \land\\
C_r \leq -1.15 \cdot Cb + 301.75 \quad \land\\
Cr \leq -2.2857 \cdot Cb + 432.85 \quad \land\\
\end{aligned}
\end{equation} 
Criterion 3:
\begin{equation}
H < 25 \quad \lor \quad H > 230
\end{equation}
Here, $\land$ denotes the logical AND operation, and $\lor$ denotes the logical OR operation. An example output of this skin detection algorithm is shown in Fig. \ref{fig: skindetection}. It can be seen from Fig. \ref{fig: skindetection} that although some non-skin pixels are picked up by the segmentation mask, most of the skin pixels are correctly identified. Note that we will find the average value within the skin region, as discussed later, so a few non-skin pixels will not significantly degrade the estimation. It can also be seen from Fig. \ref{fig: skindetection} that the algorithm can effectively handle various skin complexions across different ethnicities.

\subsection{Gray Balancing and Polynomial Transformation}\label{sec: calibration}
Image calibration is a process that converts device-dependent $RGB$ values into CIE $XYZ$ values. So we need to obtain the $RGB$ values of the target color patches in the original image as well as their corresponding CIE $XYZ$ values. The patches of interest are labelled $1$ through $35$ in Fig. \ref{fig: cc}. The CIE $XYZ$ values are measured with an X-Rite spectrophotometer under D50 illuminant, and the $RGB$ values are extracted by averaging over the center region of each patch for each color channel. 

Now that we have the ($RGB$,CIE $XYZ$) pairs, we can estimate the color mapping between them. We will utilize a two-step process \cite{gindi2008color}, namely gray balancing and polynomial regression. Gray balancing aims to remove the color cast and avoid having one particular dominant hue in the image. The methodology is based on \cite{gindi2008color}. We assume that the linear $RGB$ values of each patch and the CIE $Y$ (Luminance) value of the neutral gray patches are related by
\begin{equation}
R_l = G_l = B_l = Y.
\end{equation} 
Twelve neutral gray patches (patches No. 6 to No. 17 in Fig. \ref{fig: cc}) on the color checker are used in this step. This gives us twelve pairs of $(R_\gamma,R_l), (G_\gamma,G_l)$, and $(B_\gamma,B_l)$ for each image. We then fit a Gain-Gamma-Offset model \cite{gindi2008color} to them, such that
\begin{equation}
 \begin{aligned}
R_l &= \alpha_R \left( \frac{R_\gamma}{255}\right)^{\gamma_R} + O_R, \\
G_l &= \alpha_G \left( \frac{G_\gamma}{255}\right)^{\gamma_G} + O_G, \\
B_l &= \alpha_B \left( \frac{B_\gamma}{255}\right)^{\gamma_B} + O_B, \\
\end{aligned}
\end{equation} 
where \(\alpha_i, \gamma_i, \) and $O_i$ $(i = R,G,B)$ are the gain, gamma, and offset. The gray balancing curves of a sample image are shown in Fig. \ref{fig: graycurves}. For this particular image, $\alpha_R = 95.69, \gamma_R = 2.00, O_R = 3.48, \alpha_G = 97.34, \gamma_G = 1.93, O_G = 3.05, \alpha_B = 99.99, \gamma_B = 1.98,$ and $O_B = 3.37.$

\begin{figure}[t]
\centering
  \centerline{\includegraphics[width=0.49\textwidth]{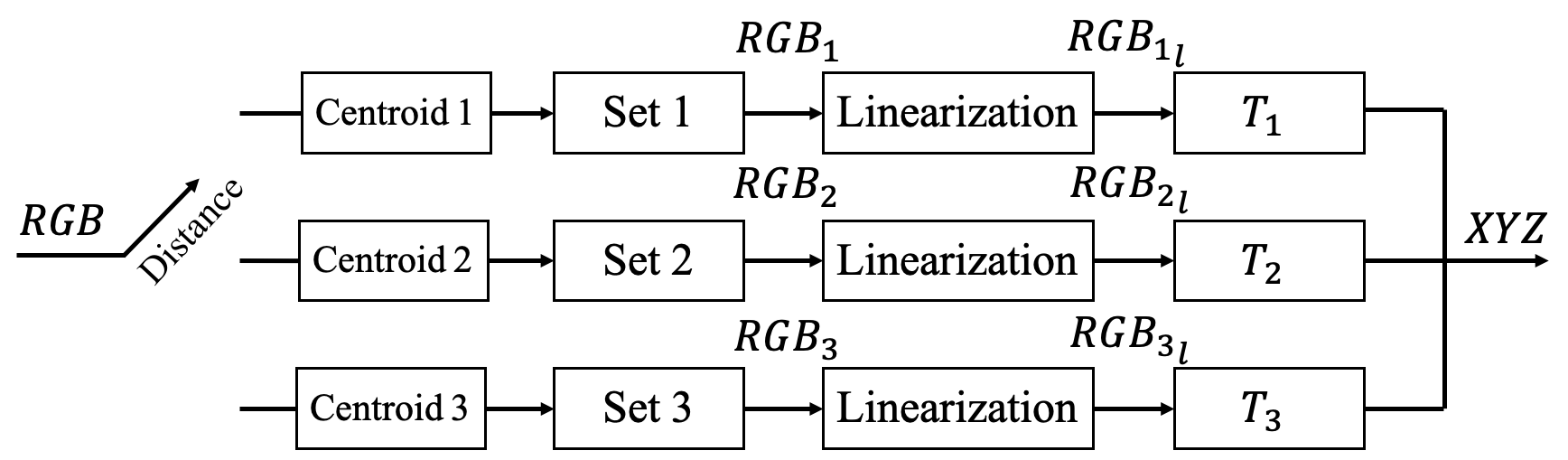}}
  \caption{Block diagram of the calibration procedure.}
  \label{fig: block}
\end{figure} 

After obtaining the linearized $RGB$ values, we can continue to the regression step. To maximize calibration accuracy, the target patches are classified into three sets, and the calibration mapping is trained separately on each set. We refer to these three training sets as Set 1, Set 2, and Set 3, respectively. We use the following procedure for patch grouping. To form Set 1, we first compute the mean $RGB$ values of the pixels within the skin segmentation mask described in the previous section. We refer to this mean as the centroid of Set 1. Then, we find the patches for which the Euclidean distance is less than 80 to the centroid of Set 1. Similarly, the patches in Sets 2 and 3 are grouped based on the centroids determined by the K-means algorithm. The centroids of these three training sets will be used again when classifying image pixels for the entire image, as discussed later. Finally, we compute three different transformation matrices \(T_i\) using polynomial regression.
Let \(Q\) denote the \(N_i \times 11\) matrix consisting of the polynomial terms of the linearized $RGB$ values of the target patches. 
\begin{equation}
\begin{split}
Q_i &= [\begin{matrix}R_{i,l} & G_{i,l} & B_{i,l} & R_{i,l}^2 & G_{i,l}^2 & B_{i,l}^2  \end{matrix}\\ 
&\qquad \quad \begin{matrix} R_{i,l}G_{i,l} & G_{i,l}B_{i,l} & R_{i,l}B_{i,l} & R_{i,l}G_{i,l}B_{i,l} & 1\end{matrix}],
\end{split}
\end{equation} where $N_i$ indicates the number of patches in the $i$-th set.
Let \(P\) denote the \(N_i \times 3\) matrix consisting of the measured CIE $XYZ$ values
\begin{equation}
P_i = [\begin{matrix} X_{i} & Y_{i} & Z_{i} \end{matrix}].
\end{equation}
Then, the optimal $3 \times 11$ transformation matrix can be computed as \begin{equation}
T_i = (Q_i^T Q_i)^{-1} Q_i^T P_i.
\end{equation} 

Now, we are ready to apply the transformation matrices to the entire image. As illustrated by Fig. \ref{fig: block}, the image pixels will be classified into three sets based on their distance to the three pre-computed centroids, and then the linearized $RGB$ values will be converted to CIE $XYZ$ using the corresponding matrices.  

\begin{figure}[t]
  \centering
\begin{subfigure}[b]{0.3\textwidth}
  \centering
  \includegraphics[width=\textwidth]{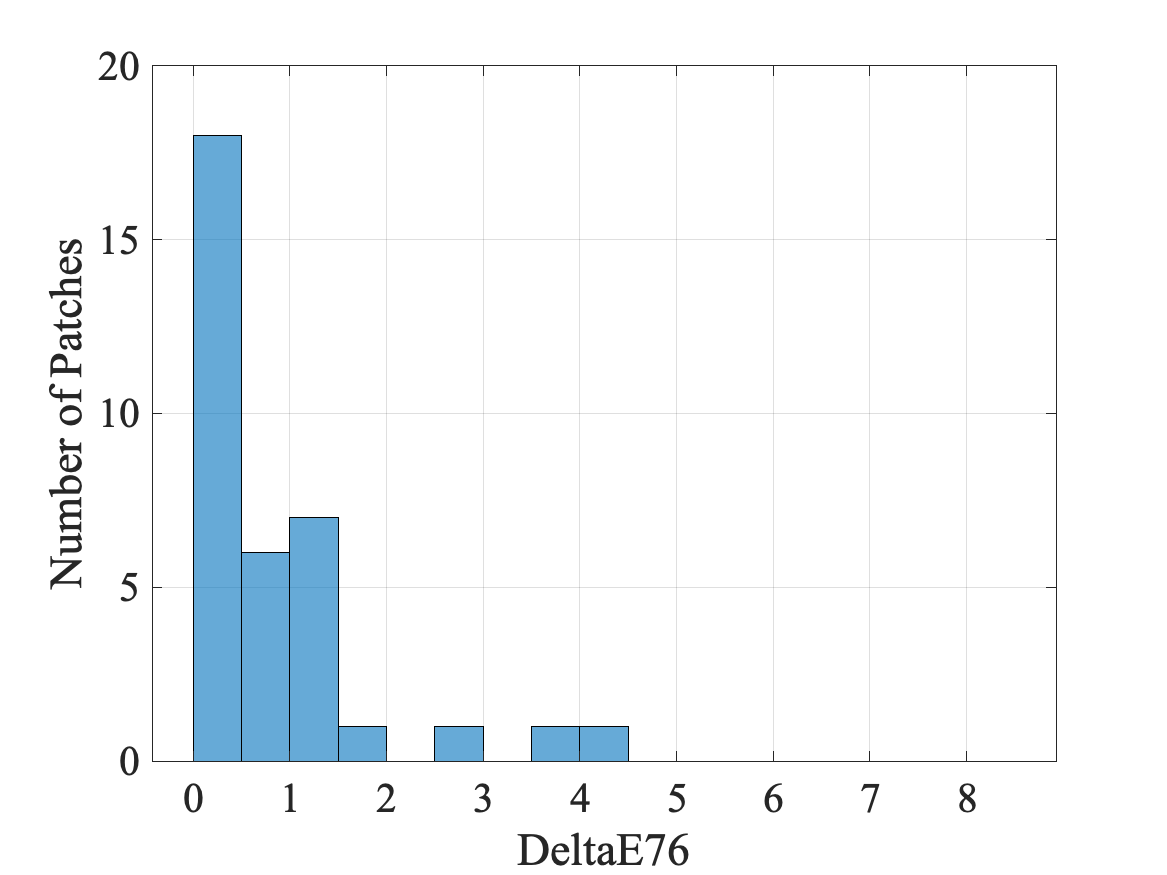}
  \centering
  \caption{\label{subfig: fig1}}
\end{subfigure}
\begin{subfigure}[b]{0.3\textwidth}
  \centering
  \includegraphics[width=\textwidth]{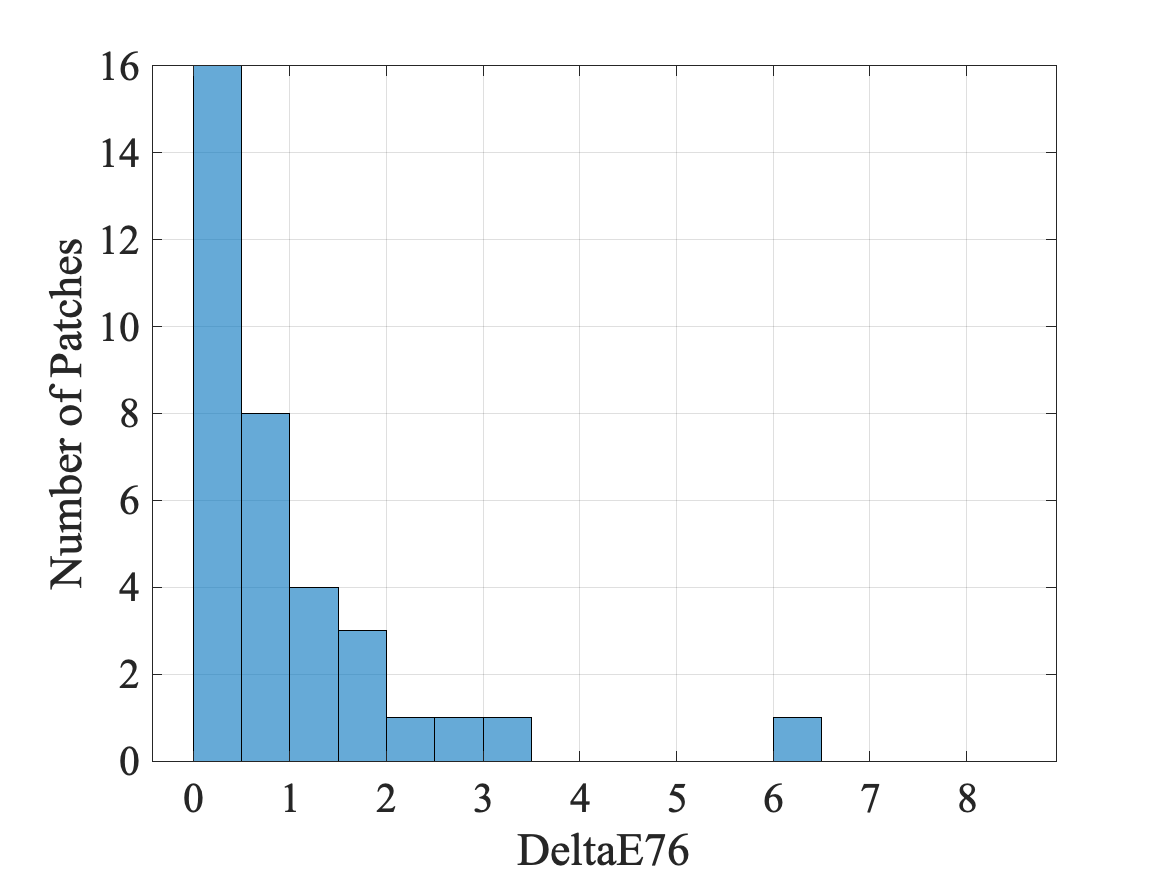}
  \centering
  \caption{\label{subfig: fig2}}
\end{subfigure}
\begin{subfigure}[b]{0.3\textwidth}
  \centering
  \includegraphics[width=\textwidth]{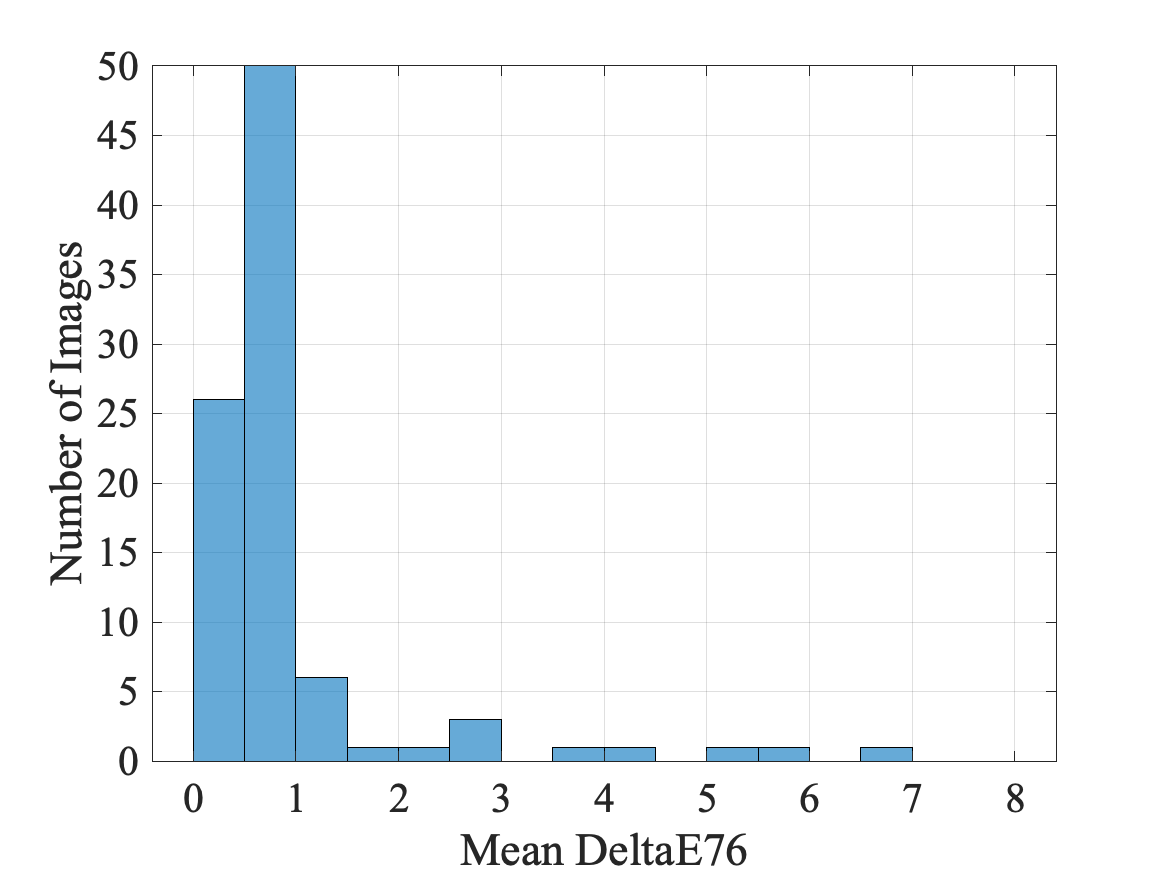}
  \centering
  \caption{\label{subfig: fig2}}
\end{subfigure}
\caption{ (a) The histogram of the calibration $\Delta E76$ across all 35 patches for the skin with no foundation image of subject No. 1 and (b) subject No. 2. The mean calibration $\Delta E76$ values over all 35 patches for subject No. 1 and subject No. 2 are 0.75 and 0.91, respectively. (c) The histogram of the mean calibration $\Delta E76$ values across all images of all subjects. }
\label{fig: calibration}
\end{figure}

\subsection{Calibration Performance}\label{sec: caliperf}
The color correction stage is now completed, and we can evaluate the accuracy of it. To better perceive the color coordinates, they are converted from CIE $XYZ$ to the 1976 CIE $L^*a^*b^*$ space.  
The conversion is defined as
\begin{equation}
\begin{aligned}
L^* &= 116 \cdot f\left( \frac{Y}{Y_n} \right) - 16 \\
a^* &= 500 \cdot \left( f\left( \frac{X}{X_n} \right) - f\left( \frac{Y}{Y_n} \right) \right) \\
b^* &= 200 \cdot  \left( f\left( \frac{Y}{Y_n} \right) - f\left( \frac{Z}{Z_n} \right) \right),
\end{aligned}
\end{equation}
where
\begin{equation}
f(t) = \begin{cases}
      \frac{1}{3} \Big( \frac{116}{24}\Big)^2 + \frac{16}{116}
, & \text{if} \quad x \leq \frac{24}{116},\\
       x^{\frac{1}{3}}
, & \text{if} \quad x > \frac{24}{116}.\\
    \end{cases}
\end{equation} Here, \((X_n, Y_n, Z_n)\) are the CIE $XYZ$ tristimulus values of the white point. With CIE standard D50 illumination, \(X_n = 96.42, Y_n = 100, and Z_n = 82.52.\)

Then, the color difference between the ground truth and the calibrated CIE $L^*a^*b^*$ coordinates of each of the patches is computed. The color difference is defined as the Euclidean distance between the color coordinates $(L_1^*, a_1^*, b_1^*)$ and $(L_2^*, a_2^*, b_2^*)$, such that
\begin{equation}
\Delta E76 = \sqrt{(L_1^* - L_2 ^ *)^2 + (a_1^* - a_2 ^ *)^2 + (b_1^* - b_2 ^ *)^2}.
\end{equation}
The resulting histograms for two sample images are shown in Figs. \ref{fig: calibration} (a) and (b). As can be seen from Figs. \ref{fig: calibration} (a) and (b), out of 35 patches, 33 patches have an $\Delta E76$ value less than 3 for both cases. So for these two images, color correction is quite effective. Figure \ref{fig: calibration} (c) shows the histogram of the mean $\Delta E76$ value over 35 patches for all images. It can be seen that the mean calibration error for the majority of the images is too small to be noticed, i.e. $\Delta E76 \leq 1$. This suggests that the proposed calibration algorithm works well on the dataset. The five images with a mean $\Delta E76$ value greater than 3 are considered to be outliers, and are therefore excluded from the subsequent analysis.

\begin{figure}[t]
\centering
\centerline{\includegraphics[width=0.49\textwidth]{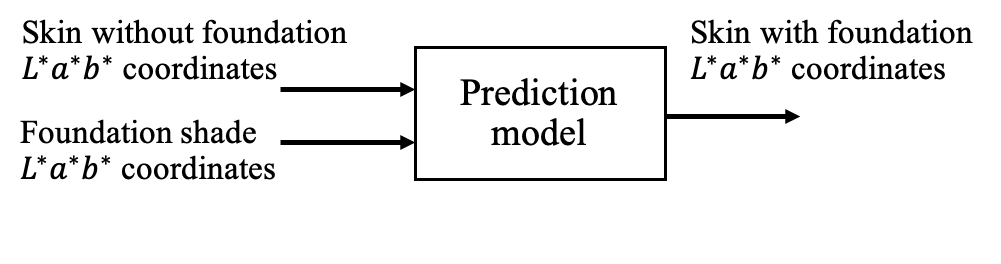}}
  \caption{Illustration of the input and output of the prediction model. }
  \label{fig: main}
\end{figure} 

\section{Experimental Results}
\begin{figure}[t]
\centering
  \centerline{\includegraphics[width=0.2\textwidth]{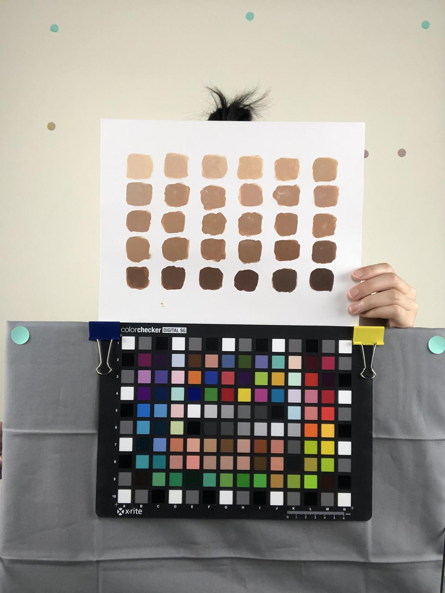}}
  \caption{The original foundation swatches image. From top to bottom and from left to right, the shades are No. 100, 110, 120, 130, 140, 150, No. 200, 210, 220, 230, 240, 250, 300, 310, ... ..., 540, 550. }
  \label{fig: foundations}
\end{figure}

With the outliers taken out, the remaining images consist of 63 skin with and without foundation pairs. These images are taken by 19 subjects. The distribution of the skin tone types is summarized in Table \ref{tbl: data}. 
\begin{table}[h]
 \centering
\setlength\tabcolsep{4pt}
\begin{tabular}{|c|c|c|c|c|c|}
\hline
Skin Tone Type        & Fair & Light & Medium & Tan & Dark \\ \hline
Number of Subjects &    4  &    4   &    4    &  3   &    3  \\ \hline
Number of Image Pairs &    14  &    13   &    14    &  12   &    10  \\ \hline
\end{tabular}
\caption{Number of subjects and number of images in each skin tone category. }
\label{tbl: data}
\end{table} After color correcting all the images using our proposed method, we obtain 63 pairs of CIE $L^*a^*b^*$ coordinates.
Apart from the selfie photos of the subjects, we also collect an image that contains the swatches of all the foundation shades. The foundation shades are applied on a white cardboard using makeup sponges. The original foundation swatches image is shown in Fig. \ref{fig: foundations}. The same color correction procedure is performed on the image, except that the three sets are now all determined by K-means. The mean calibration $\Delta E76$ of this foundation swatches image is 0.34. 

Now, we can develop a model to predict the CIE $L^*a^*b^*$ coordinates of the skin-with-foundation color, given the CIE $L^*a^*b^*$ coordinates of a skin-with-no-foundation color and the CIE $L^*a^*b^*$ coordinates of a foundation color, as illustrated in Fig. \ref{fig: main}. We explored two common machine learning models. They are linear regression and SVR \cite{drucker1996support} with a linear kernel. 

We denote the 6-dimensional input vector (i.e., the CIE $L^*a^*b^*$ coordinates of a skin-with-no-foundation color and the CIE $L^*a^*b^*$ coordinates of a foundation color) of the $i$-th sample as $x_i$ and the corresponding 3-dimensional ground truth label (i.e., the CIE $L^*a^*b^*$ coordinates of the skin-with-foundation color) as $y_i$. Then, the linear regression problem can be formulated as
\begin{equation}
J(\theta) = \sum_{i=1}^N \mathcal{L}(\theta^T x_i, y_i) = || \theta^T x_i - y_i ||^2,
\label{eqn: lrcost}
\end{equation}
where \(\theta = [w, b]^T\), \(\theta^T x_i\) is the prediction made by the model, and $N$ is the number of samples in the dataset. We further define a $N \times 7$ matrix
\begin{equation}
    A = \begin{pmatrix}
x_1^T & 1\\
x_2^T & 1\\
... & ...\\
x_N ^T & 1
\end{pmatrix}, 
\end{equation} and a $N \times 3$ matrix
\begin{equation}
    Y = \begin{pmatrix}
y_1^T\\
y_2^T\\
... \\
y_N ^T
\end{pmatrix}.
\end{equation} 
Then, (\ref{eqn: lrcost}) can be converted to the vector form such that \(J(\theta) = || A \theta- Y ||^2\).
Thus, the optimized solution can be expressed as
\begin{equation}
\hat{\theta} = \operatorname*{argmin}_\theta J(\theta) = (A^TA)^{-1}A^TY.
\end{equation}

The SVR algorithm is based on the same concept as the support vector machine, except that it uses the support vectors for soft margins in the regression process rather than classification. The regression in SVR can be formulated as:
\begin{equation}
\operatorname*{argmin}_{w,b} \frac{1}{2} ||w||^2 + C \sum_i^N (\xi_i + \xi_i^*),
\end{equation} 
subject to
\begin{equation}
\begin{aligned}
	y_i - wx_i - b &\leq \epsilon + \xi_i \\
	wx_i + b - y_i &\leq \epsilon + \xi_i^* \\
	\xi_i, \xi_i^* & > 0,
\end{aligned}
\end{equation}
where $C$ is the box constraint, \(\epsilon\) is the margin around the decision boundary, and \(\xi_i, \xi_i^*\) are the slack variables. With a linear kernel, the prediction function can be written as:
\begin{equation}
f(x) = \sum_{i = 1} ^ N (\alpha_i - \alpha_i^*) \langle x_i,x \rangle + b,
\end{equation} where $\alpha_i$ and $ \alpha_i^*$ are the Lagrange multipliers. The Python package \textit{Scikit-learn} is used to implement linear regression and SVR. 

A common way to evaluate the performance of a regression model is to use $K$-fold cross-validation. In $K$-fold cross-validation, the dataset is randomly split into $K$ equal-sized folds. We train the model on the data in $K-1$ of the folds and evaluate the model on the remaining one fold, namely, the validation fold. We then repeat this process $K$ times. Given the fact that we have limited data in the dataset, we choose $K = N$, where $N$ is the number of all data points. This method is referred to as the leave-one-out cross-validation (LOOCV) method. That is to say, in each trial, the predictor is trained on all but one data point, and the prediction is made for that excluded point. Then the performance can be computed as the average over the $N$ trials. The advantage of using LOOCV is that each data point gets the chance to be allocated into both the testing set and $N-1$ of the training sets. The selection bias is therefore decreased. 

To determine the goodness of model fit, the coefficient of determination, denoted by $R^2$, is computed. In the context of regression, it is a measure of the proportion of the prediction error that can be attributed to the variance in the independent input variables. It is defined as:
\begin{equation}
R^2 = 1 - \frac{\sum_{i = 1} ^ N (y_i - \hat{y}_i)^2}{\sum_{i = 1} ^ N (y_i - \bar{y})^2},
\end{equation} 
where \(y_i - \hat{y}_i\) is the error from the prediction \(\hat{y}_i\) to the ground truth \(y_i\) of the \(i\)-th data point, and \(\bar{y}\) is the mean of the ground truth \(y\) values. The $R^2$ score is in the range of \([0,1]\). A score of 0 means that the dependent variable cannot be predicted from the independent variable, and a score of 1 means the dependent variable can be predicted with no error from the independent variable.
 
Besides, the mean squared error (MSE) and mean absolute error (MAE) are also used to evaluate the accuracy of prediction results. The MSE can be expressed as
 \begin{equation}
 MSE = \frac{\sum_{i = 1} ^ n (p_i - g_i)^2}{n},
 \end{equation} where $p_i$ and $g_i$ are the predicted value and the ground truth value of the $i$-th sample point, respectively.
Similarly, 
 \begin{equation}
 MAE = \frac{\sum_{i = 1} ^ n |p_i - g_i|}{n}.
 \end{equation} 
 
Table \ref{tbl: results} summarizes the LOOCV results of the linear regression model and the SVR model in terms of $R^2$, average MSE, and average MAE. It can be seen that the average MSE and MAE values are less than 1.5 $\Delta E$ and the $R^2$ value is high for both of the models. This implies that the prediction models can accurately predict the skin with foundation color on the dataset.
 \begin{table}[h]
 \setlength\tabcolsep{2.7pt}
 \centering
\begin{tabular}{|c|c|c|c|}
\hline
Model                  & $R^2$    & Average MSE  & Average MAE  \\ \hline
Linear Regression      & 0.83 & 1.50  & 0.91  \\ \hline
SVR with a Linear Kernel & 0.82 & 1.49  & 0.87 \\ \hline
\end{tabular}
\caption{LOOCV $R^2$, MSE, and MAE results.}
\label{tbl: results}
\end{table}

 \section{Conclusion}
The selfie images are calibrated using a subset of color checker patches. The pixels are classified into three sets according to the Euclidean distance from the $RGB$ values of the pixel to the three designated centroids. Three different transformation matrices are computed separately and then applied to the corresponding pixels in the image. The calibration accuracy is measured by the color difference $\Delta E$ between the reference value and the calibrated value in CIE $L^*a^*b^*$ space. The $\Delta E$ results indicate that the error produced by the proposed method is almost not distinguishable for most of the images. A prediction model is then built upon the calibrated selfie images. The prediction performance is measured by $R^2$, $MSE$, and $MEA$. LOOCV results show that the prediction made by both linear regression and SVR with a linear kernel is reliable.




\bibliographystyle{IEEEbib}
\bibliography{mybibfile}


\begin{biography}
%
Yafei Mao received her BS in Electrical and Computer Engineering from Purdue University (2016). Since then, she has been working on her PhD in Electrical and Computer Engineering at Purdue University. Her research interests are haltoning, image processing, and computer vision.



\end{biography}

\end{document}